%% file: main.tex
\documentclass{article}
\usepackage{iclr2027_conference,times}

\input{math_commands.tex}

\usepackage{hyperref}
\usepackage{url}
\usepackage{amsmath}
\usepackage{amssymb}
\usepackage{booktabs}
\usepackage{multirow}
\usepackage{graphicx}
\usepackage[dvipsnames]{xcolor}
\usepackage{algorithm}
\usepackage{algorithmic}
\usepackage{wrapfig}
\usepackage[most]{tcolorbox}
\usepackage{needspace}
\usepackage{placeins}
\usepackage{capt-of}

\definecolor{cneu}{HTML}{5F6B7A}
\definecolor{cok}{HTML}{2E7D32}
\definecolor{cbad}{HTML}{B3261E}
\newtcolorbox{solbox}[2][]{enhanced, breakable,
  colback=#2!4!white, colframe=#2!80!black, coltitle=white,
  fonttitle=\small\bfseries, fontupper=\small, title={#1},
  boxrule=0.6pt, arc=1.2mm, left=5pt, right=5pt, top=3pt, bottom=3pt,
  toptitle=1.5pt, bottomtitle=1.5pt, before skip=5pt, after skip=5pt}

\newcommand{\method}{\textsc{PRISM}}
\newcommand{\argd}{\textsc{ARG-Designer}}

\title{PRISM: \textbf{PRI}vileged \textbf{S}ibling-guided \\
\textbf{M}ulti-Agent Topology Design}

\author{Phuong Tuan Dat$^{1}$, Qi Li$^{1}$, Xinchao Wang$^{1,\dagger}$ \\
$^{1}$National University of Singapore \\
\{phuongtuandat, liqi\}@u.nus.edu, xinchao@nus.edu.sg \\
$^{\dagger}$Corresponding author}

\iclrfinalcopy
\begin{document}

\maketitle
\lhead{Preprint. Under review.}

\begin{abstract}
Multi-agent systems (MAS) based on large language models (LLMs) have become a powerful way to address complex problems, and their effectiveness depends critically on the collaboration topology that decides which agents participate and how they communicate. Recent work learns to generate this topology from the task query, fitting the generator to corpora of execution-verified graphs. The post-training method for such generators remains a limitation: a corpus lists the graphs that solved a query but never ranks them, leaving continued supervised fine-tuning (SFT) with no per-query preference to exploit. Without such a preference, the generator drifts toward the topologies the corpus contains most often rather than those that suit each query, so a gain on one task can come at a cost on another. We propose \method{}, which recovers per-query signal from the same corpus by learning with privileged information: a teacher reads, through a zero-initialized low-rank branch, a sibling graph that also solved the query and is unavailable at deployment, and the student matches its per-decision targets on trajectories the student samples itself. Across six diverse benchmarks \method{} improves over every topology-design baseline evaluated, and matched-teacher controls support the contribution of privileged query information. Code is available in \href{https://github.com/tuandattt/PRISM.git}{this link}.
\end{abstract}

\section{Introduction}

Multi-agent Systems based on Large Language Models (LLMs) address complex tasks by coordinating agents with complementary roles. Their effectiveness depends on the collaboration topology: the graph specifying which agents participate and how information flows between them \citep{zhuge2024gptswarm,shen2025propagation}. Consequently, improving MAS requires learning how to organize collaboration alongside improving the capabilities of individual agents. Research on automated topology design has expanded from predefined workflows to learned communication structures and, more recently, generators that construct a collaboration graph for each query \citep{li2026argdesigner, zhang2025gdesigner}. These generators learn which agents to include and how to connect them for a given query.

Recent work has made query-conditioned topology generation feasible by deriving supervision from multi-agent execution outcomes \citep{li2026argdesigner, ofa-mas}. Candidate collaboration graphs are instantiated and executed on training queries, and empirically successful query–graph pairs are retained as training examples. Using these verified pairs, an autoregressive topology generator learns through teacher forcing to predict agent roles and communication links conditioned on the query and the partially constructed graph. This approach turns previous collaboration outcomes into reusable supervision and provides a straightforward post-training strategy: continuing supervised fine-tuning (SFT) on verified graphs. However, closer agreement with the recorded graphs does not necessarily translate into better downstream task performance.

In our experiments, continued SFT concentrates the generated topology distribution, while its effects on downstream performance vary across benchmarks: some improve, whereas others exhibit limited gains or regressions. This observation is consistent with prior findings that likelihood-based SFT can reduce output diversity \citep{preserving_diversity}, although concentration alone does not establish whether the resulting topologies are more effective. Teacher forcing also introduces a mismatch between the partial graphs encountered during training and inference. Training supervises decisions along recorded construction trajectories, whereas deployment requires the generator to act on partial graphs produced by its own preceding decisions. Deviations from recorded trajectories can therefore lead to partial graphs for which the verified corpus provides no direct supervision. This mismatch parallels exposure bias in autoregressive imitation learning and motivates on-policy distillation \citep{autoregressive_KD, agarwal2024gkd}: using a teacher to guide construction decisions at partial graphs visited by the student.

In this work, we propose \method{}, which addresses this training-inference mismatch by distilling verified-graph information at the partial graphs actually visited by the student. For each training query, the student first generates its own partial graph trajectory. A privileged teacher then receives the query together with another verified graph for the same query and provides a soft distribution over the next topology decisions at each student-generated partial graph. Thus, supervision remains available even after the student deviates from the recorded verified trajectory, directly addressing the off-trajectory states left unsupervised by teacher-forced SFT. The query-matched verified graph also provides information about a successful alternative construction, while supervised anchoring preserves the verified graph trajectories and discourages the student from drifting away from valid topology patterns. Since the verified graphs are collected before post-training, \method{} requires no additional MAS executions, and the privileged input is removed at inference time without changing the deployed topology generator.

In summary, this work makes the following contributions:
\begin{itemize}
\itemsep2pt \parskip0pt
\item We document increased output concentration under continued supervised fine-tuning of an autoregressive topology designer and relate it to accuracy across benchmarks.

\item We propose \method{}, which uses successful sibling graphs as training-only privileged inputs for on-policy self-distillation, without additional post-training LLM executions or changes to the deployed designer architecture.

\item We demonstrate its effect across six benchmarks spanning mathematical reasoning, general reasoning and code generation - GSM8K, MultiArith, SVAMP, AQuA, MMLU and HumanEval - outperforming the strongest topology-generation baseline on each. Matched teacher controls support an additional benefit from query-matched graph information beyond constant-input on-policy distillation.
\end{itemize}

\section{Related Work}
\label{sec:related}

\paragraph{MAS communication topology design.}
Early approaches organize reasoning and collaboration through chains \citep{wei2022cot,hong2024metagpt}, tree-structured search \citep{yao2023tot}, and debate protocols \citep{du2023debate}. In MAS, performance depends on the communication structure \citep{qian2025scaling,shen2025propagation}, while the computational cost of collaboration has motivated research on efficiency \citep{MAS_survey}. Automated topology design methods optimize agent participation and communication: AgentPrune \citep{zhang2025agentprune} prunes communication edges, AgentDropout \citep{wang2025agentdropout} removes agents and edges across communication rounds, and G-Designer \citep{zhang2025gdesigner} generates task-conditioned communication graphs using a variational graph autoencoder. These methods operate on a predefined set of candidate agents. \argd{} \citep{li2026argdesigner} extends the design space by autoregressively generating agent roles and communication edges, allowing the number of agents and their role composition to vary with the query. Complementary research investigates on-policy refinement using failure trajectories for agent safety alignment \citep{yin2026policy} and adversarial manipulation that increases the computational cost of tool-augmented reasoning \citep{li2026sponge}. Our work focuses on post-training an autoregressive topology generator initialized through supervised learning on execution-verified graphs. We retain \argd{}'s deployed architecture and introduce a privileged teacher to provide additional supervision along student-generated graph trajectories.

\paragraph{On-policy distillation.} DAgger \citep{ross2011dagger} learns from states the policy itself visits rather than states a demonstrator recorded, and GKD \citep{agarwal2024gkd} and MiniLLM \citep{gu2024minillm} bring that idea into LLM distillation, sampling from the student and scoring with a teacher; the latter also motivates mode-seeking reverse KL. A further line makes teacher and student one model rather than a large-small pair: \citet{zhao2026self} distil a reasoner from its own on-policy samples, and \citet{dat2026dopsd} carry the scheme to diffusion language models. \method{} uses on-policy self-distillation in this sense, separating teacher from student by a privileged input rather than by scale or by an earlier checkpoint. Its decision spaces are also small categorical decision spaces (roles $+$ \texttt{END}; edge/no-edge), so the divergence is exact rather than estimated.

\section{Preliminaries}
\label{sec:prelim}

\paragraph{MAS as a collaboration graph.} Following \citet{li2026argdesigner}, a MAS is a directed acyclic graph $G=(\mathcal{V},\mathcal{E})$ whose nodes are LLM instances carrying roles from an extensible pool $\mathcal{R}$, and where an edge $e_{ji}=(v_j,v_i)$ means $v_i$ receives $v_j$'s output in its prompt. Agents execute in topological order for $K$ rounds and a summariser aggregates the final responses. Each agent is queried once per round and the summariser once per question, so executing $G$ costs $K|\mathcal{V}| + 1$ LLM calls.

\paragraph{Conditional autoregressive generation.} The designer factorises the conditional distribution over graphs as
\begin{equation}
P_\theta(G \mid Q, \mathcal{R}) \;=\;
\prod_{i=1}^{|\mathcal{V}|}
\underbrace{P_\theta(v_i \mid G_{<i}, Q, \mathcal{R})}_{\text{node generation}}
\cdot
\prod_{j=1}^{i-1}
\underbrace{P_\theta(e_{ji} \mid v_i, G_{<i}, Q)}_{\text{edge generation}} ,
\label{eq:factorisation}
\end{equation}
where $G_{<i}$ is the subgraph over the first $i-1$ nodes. A sentence encoder maps $Q$ to $f_Q \in \mathbb{R}^d$ and a task processor produces the conditioning vector $c = \mathrm{TaskProc}(f_Q)$, gated with a GRU summary of the previously selected roles \citep{cho2014gru}. A node GRU consumes that context and a metric-learning head scores roles by inner product against role embeddings extended with a learnable \texttt{END} token - the mechanism that keeps the role pool extensible - after which an edge GRU emits the $i-1$ incoming-edge decisions for $v_i$. Training minimises the negative log-likelihood, decomposed as
\begin{equation}
\mathcal{L}_{\text{total}} \;=\;
\alpha \, \mathcal{L}_{\text{node}} + (1-\alpha)\, \mathcal{L}_{\text{edge}}
\label{eq:nll}
\end{equation}
under teacher forcing.

\paragraph{The verified corpus.} The supervision set is constructed without additional human annotation: for each training query $Q$, blueprint configurations $c \in \mathcal{C}$ (chain, star, layered, fully-connected, mesh, with random roles and agent counts) are instantiated as graphs $G(c)$, executed, and retained only if they solved the query,
\begin{equation}
\mathcal{D} \;=\; \bigl\{\,(G(c), Q) \;\big|\; S\bigl(Q, G(c)\bigr) = 1 \,\bigr\},
\qquad S(\cdot) \in \{0,1\}.
\label{eq:corpus}
\end{equation}
One such corpus is built for each evaluated domain; their sizes, blueprint composition and query-level splits are reported in Appendices~\ref{app:corpus} and~\ref{app:splits}. What matters here is the shape of $\mathcal{D}$ rather than its scale: a query is typically solved by several distinct topologies, so the corpus attaches to each query a set of graphs that worked and never marks one of them as preferable.\label{sec:premise}

\section{PRISM}
\label{sec:method}

\begin{figure}[t]
\begin{center}
\IfFileExists{figures/OPSDesigner.pdf}
{\includegraphics[width=\linewidth,trim={18 13 14 12},clip]{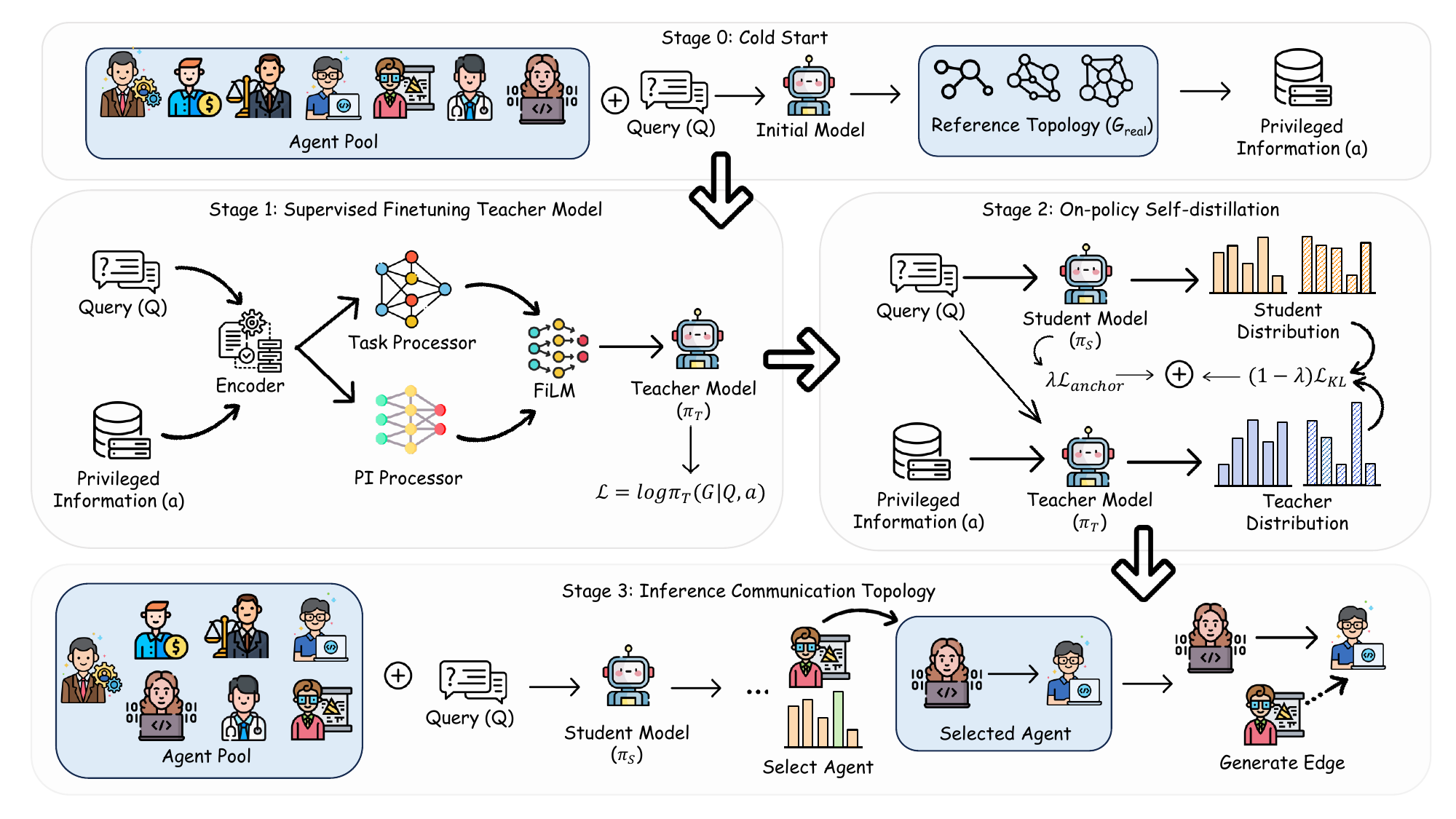}}
{\fbox{\rule[-1.2cm]{0cm}{2.4cm}\rule[-.5cm]{0.46\linewidth}{0cm}}}
\end{center}
\vspace{-10pt}
\caption{Overview of \method{}. \textbf{Stage 0} fits the designer on execution-verified topologies, which also supply the privileged input $a$. \textbf{Stage 1} trains a separate teacher whose FiLM branch reads $a$. \textbf{Stage 2} has the teacher rescore the student's own trajectories, the student minimising Eq.~\ref{eq:opsd}. \textbf{Stage 3} deploys the student alone; the teacher is discarded.}
\vspace{-4mm}
\label{fig:method}
\end{figure}

\method{} provides an alternative supervision pathway for exploiting the query--graph associations already present in the verified corpus. As shown in Figure \ref{fig:method}, Stage~1 constructs a teacher conditioned on a successful sibling graph, and Stage~2 distils its predictions on trajectories sampled from the student's current policy.

\subsection{Stage 1: Construct the privileged teacher}

Let $a$ be privileged information for query $Q$: information that exists at training time and is structurally unavailable at deployment. We clone the student and give the clone one extra input branch; writing $f_a$ for the encoding of $a$, the teacher's conditioning vector is a FiLM modulation \citep{perez2018film} of the student's:
\begin{equation}
c_{\text{teacher}} \;=\; \mathrm{TaskProc}(f_Q) \odot \bigl(1 + \gamma(f_a)\bigr) \;+\; \beta(f_a),
\qquad
\gamma, \beta : \mathbb{R}^{r} \to \mathbb{R}^{d},
\label{eq:film}
\end{equation}
where $f_a$ passes through a small encoder - layer normalisation, dropout, a linear map to rank $r=16$ and a $\tanh$ activation - before $\gamma$ and $\beta$ are computed. Starting from pretrained ARG-Designer checkpoint $\theta_0$, the teacher minimizes the weighted supervised objective in Eq.~\ref{eq:nll}, updating all teacher parameters. The student itself receives no updates and remains at $\theta_0$ throughout this stage.

\paragraph{Sibling sampling and encoding.} Each successful graph is converted into a deterministic text representation of its roles and edges, which is embedded using the same frozen MiniLM encoder \citep{wang2020minilm} used for the query. Duplicate graph representations are merged before sibling sampling. For a verified target graph $G$, the sibling is drawn uniformly from the other graphs that solved the same query, skipping any whose text representation is identical to $G$'s, which keeps the target itself out of the privileged channel.
 The sibling is resampled every epoch, allowing the pairing to change when multiple eligible candidates exist.

If a query has no distinct successful sibling, its privileged input is a zero vector; no graph from another query is substituted. This supplies no sibling-specific graph information, although a zero input does not necessarily produce an identity FiLM modulation after training. The implementation counts queries without siblings and stops training if no query has an eligible sibling.

The teacher is a separate model initialized from a copy of the student weights $\theta_0$ and augmented with a privileged encoder and FiLM branch; no parameters are shared between the models. Under identical deterministic evaluation conditions, zero initialization of the FiLM output maps makes the teacher's predictions identical to the student's at teacher creation. This removes an initial functional mismatch, but does not imply equal model capacity or zero divergence after teacher training. During Stage~2 the teacher is frozen and every student parameter is trainable. The channel is kept narrow because the cold start yields only $O(10^3)$ graphs: at rank $16$ it adds ${\approx}20$k parameters against ${\approx}592$k for a full-width channel at $d=384$, an approximately 30-fold reduction in channel parameters while retaining competitive accuracy in the evaluated settings (\S\ref{sec:ablation}).

\subsection{Stage 2: on-policy self-distillation}

At the start of Stage~2, we sample one verified reference graph uniformly from the deduplicated set of successful topologies for each training query. This reference is fixed throughout Stage~2 and shared across all $M$ rollouts for that query. Unlike the target-exclusion rule used in Stage~1, Stage~2 does not exclude references that match the supervised anchor or a sampled rollout. Queries with a single verified topology use that topology as their reference. Student trajectories are resampled at every optimization step.

For each query we sample $M$ complete topologies from the current student policy at temperature $\tau$, so that the corrected states are always the ones this policy visits, unlike off-policy pseudo-labelling, where graphs are generated once and reused as the student drifts away from them. Each trajectory is then re-scored by both models under teacher forcing on that trajectory: the student without $a$, the teacher with $a$. Although their parameters are separate, the teacher and student predict over the same node and edge symbol sets, so every per-decision categorical divergence between them is computed exactly rather than estimated:
\begin{equation}
\mathcal{L}_{\mathrm{KL}} \;=\;
\alpha \cdot \mathbb{E}_{\text{node steps}}\!\left[ \mathrm{KL}\bigl(P_\theta \,\|\, P_{\text{teacher}}\bigr) \right]
\;+\;
(1-\alpha) \cdot \mathbb{E}_{\text{edge slots}}\!\left[ \mathrm{KL}\bigl(P_\theta \,\|\, P_{\text{teacher}}\bigr) \right],
\label{eq:kl}
\end{equation}
with $\alpha=0.2$ inherited from Eq.~\ref{eq:nll}. We use reverse KL to encourage concentration on decisions favored by the teacher and evaluate this choice against alternative divergences \citep{gu2024minillm}. To retain direct supervision from verified query–graph pairs, we combine the distillation objective with a supervised anchoring loss:
\begin{equation}
\mathcal{L}_{\text{anchor}} \;=\; -\log P_\theta(G_{\text{real}} \mid Q),
\qquad
\mathcal{L}_{\method{}} \;=\; (1-\lambda)\,\mathcal{L}_{\mathrm{KL}}
\;+\; \lambda\,\mathcal{L}_{\text{anchor}}
\label{eq:opsd}
\end{equation}
where $G_{\text{real}}$ is a verified graph for $Q$ drawn from the corpus of Eq.~\ref{eq:corpus}. The two terms are normalised differently and we keep the convention explicit: $\mathcal{L}_{\text{anchor}}$ sums the log-probabilities of all node and edge decisions of $G_{\text{real}}$, unweighted by $\alpha$, and averages over the batch, whereas each divergence in Eq.~\ref{eq:kl} is a mean over decision points of its own kind. Setting $\lambda = 1$ leaves the anchor alone, which is the objective of our SFT control, making it the far endpoint of this one-parameter family rather than a different method. Rollouts are scored as sampled, never repaired first: a repaired graph is not what the policy emitted, and its gradient would not match the state it came from. Gradients flow only through the re-scoring of those trajectories; sampling itself is not differentiated and the teacher is evaluated without gradients, so we optimise the divergence at the visited states rather than an exact gradient of the expectation over $\pi_\theta$, as is standard in on-policy distillation \citep{agarwal2024gkd}. The teacher is frozen throughout Stage~2 and discarded before evaluation, so the deployed model is architecturally identical to \argd{} and the privileged channel exists only during training. The full loop is Algorithm~\ref{alg:opsd} in Appendix~\ref{app:algo}.

\section{Experiments}
\label{sec:experiments}

\begin{table}[t]
\caption{Pass@1/accuracy (\%) on six benchmarks with Qwen2.5-7B-Instruct.
Best results are bold. On AQuA, SFT trails \argd{} by $12.60\%$, while
\method{} improves over it.}
\label{tab:main}
\begin{center}
\small
\begin{tabular}{lcccccc}
\toprule
Method                   & GSM8K & MultiArith & SVAMP & AQuA & HumanEval & MMLU \\
\midrule
Vanilla (single agent)   & 70.28 & 88.33 & 82.40 & 61.81 & 69.51 & 62.70 \\
AgentPrune               & 76.72 & 94.67 & 91.90 & 68.90 & 71.95 & 67.93 \\
AgentDropout             & 77.86 & 95.33 & 92.20 & 70.47 & 77.44 & 68.32 \\
G-Designer               & 78.01 & 97.33 & 91.80 & 74.41 & 78.05 & 71.52 \\
\argd{} (baseline)       & 79.98 & 98.83 & 92.50 & 75.20 & 79.27 & 71.98 \\
SFT (control)            & 86.13 & 98.83 & 92.40 & 62.60 & 80.49 & 73.22 \\
\midrule
\method{}                & \textbf{88.02} & \textbf{99.17} & \textbf{93.30} & \textbf{76.38} & \textbf{82.32} & \textbf{73.94} \\
\bottomrule
\end{tabular}
\end{center}
\vspace{-5mm}
\end{table}

\subsection{Setup}

\textbf{Benchmarks and backbone.} Following \citet{li2026argdesigner} we evaluate on MMLU \citep{hendrycks2021mmlu}, GSM8K \citep{cobbe2021gsm8k}, MultiArith \citep{roy2016multiarith}, SVAMP \citep{patel2021svamp}, AQuA \citep{ling2017aqua}, and HumanEval \citep{chen2021humaneval}, reporting Pass@1 / accuracy. All agents run on Qwen2.5-7B-Instruct \citep{qwen2025qwen25} served locally, so every arm in every table is a controlled re-run under one execution stack rather than a number copied from a paper. The role encoder is \texttt{all-MiniLM-L6-v2} \citep{wang2020minilm} with $d=384$, and $K=1$ round with a summariser. Stage~1 uses $40$ epochs at lr $10^{-4}$; Stage~2 uses $30$ epochs at $10^{-4}$, $16$ queries per step, $M=4$ rollouts per query, $\tau=1.0$, $\alpha=0.2$, $\lambda=0.3$ and reverse KL (Appendix~\ref{app:hparams}), shared across benchmarks with no per-benchmark tuning.

\textbf{Data source and partitioning.} Our verified graph corpus uses questions from \textsc{MixChain}, MBPP and MMLU. The training and validation data are partitioned by query rather than by graph. Graphs are grouped by their associated query, the query groups are shuffled with a fixed seed, and $20\%$ are assigned to validation.
While \textsc{MixChain} supplies the corpus for the four mathematical benchmarks, general reasoning and code generation are covered separately: Stage~0 builds a verified corpus for MMLU from its own questions and, for code generation, from MBPP \citep{austin2021mbpp}, under the same procedure (Eq.~\ref{eq:corpus}); those graphs serve as the training targets and privileged siblings for the corresponding benchmark. HumanEval is therefore used for evaluation only, so no question it scores has contributed a training graph to any arm.

\textbf{Baselines and controls.} We compare against \argd{} \citep{li2026argdesigner}, G-Designer \citep{zhang2025gdesigner}, AgentPrune \citep{zhang2025agentprune} and AgentDropout \citep{wang2025agentdropout}, and a single-agent vanilla run. \argd{} is reproduced in full: the cold start of Eq.~\ref{eq:corpus} followed by the published efficiency fine-tuning stage, so the row labelled \argd{} is the complete training procedure of \citep{li2026argdesigner} rather than an initialization-only variant. The critical control is \textbf{SFT}: continued supervised maximum likelihood on the verified corpus from that same deployed checkpoint $\theta_0$, matched to Stage~2 in epochs and learning rate. Both SFT and PRISM are initialized from the deployed ARG-Designer checkpoint. Their student-training stages use 30 epochs and a learning rate of \(10^{-4}\). PRISM additionally trains a teacher for 40 epochs and performs trajectory sampling and teacher rescoring during distillation. Both methods reuse the existing verified corpus without additional multi-agent executions during post-training. These settings match the student-training schedule, but not total training computation. Table~\ref{tab:main} scores the model each procedure finally deploys. Beyond these arm-level controls, Appendix~\ref{app:intervention} executes chosen crews directly on a fixed question set, holding wiring, rounds, decision method and backbone fixed so that only the role composition varies.

\subsection{Main results}
\label{sec:main}

\method{} is the best method on every benchmark, but two features of Table~\ref{tab:main} matter more than the headline. The first is the AQuA column: SFT tracks or beats the baseline on five of the six benchmarks, by at most $6.15\%$ and never falling more than $0.10\%$ below, then drops $12.60\%$ on AQuA. The effect of continued SFT varies substantially across benchmarks. The second is the size of \method{}'s margin over SFT, which varies by an order of magnitude across those same benchmarks: $0.72\%$ on MMLU and $1.89\%$ on GSM8K against $13.78\%$ on AQuA.

\noindent
\begin{minipage}[t]{0.50\textwidth}
\vspace{0pt}
\textbf{Cost.} Table~\ref{tab:cost} reports the deployment cost of each designer. \method{} averages $3.04$ agents per query against $3.00$ for \argd{} and SFT alike. At $K=1$ that is $4.04$ LLM calls against $4.00$, buying $2.56\%$ and $3.24\%$ more accuracy for $1.0\%$ more calls. Post-training adds no executions either, since its rollouts sample graph trajectories rather than running any agent.
\end{minipage}\hfill%
\begin{minipage}[t]{0.46\textwidth}
\vspace{-11pt}
\centering
\captionof{table}{Mean per-query cost over the evaluation sets, at $K=1$.}
\label{tab:cost}
\vspace{2pt}
\small
\begin{tabular}{lccc}
\toprule
Designer & $|\mathcal{V}|$ & $|\mathcal{E}|$ & calls \\
\midrule
\argd{}   & $3.00$ & $2.30$ & $4.00$ \\
SFT       & $3.00$ & $2.21$ & $4.00$ \\
\method{} & $3.04$ & $2.27$ & $4.04$ \\
\bottomrule
\end{tabular}
\end{minipage}

\subsection{Output concentration}
\label{sec:collapse}

\begin{table}[t]
\caption{Frequency and conditional accuracy of SFT's modal role assignment, \texttt{[Math Solver, Programming Expert, Programming Expert]} on every benchmark. Share is measured in the deployed single-sample decode, so it is not comparable with the resampled $0.346$ of \S\ref{sec:collapse}.}
\label{tab:fixedcrew}
\begin{center}
\small
\begin{tabular}{lcccc}
\toprule
Benchmark & share emitted & acc.\ @ default & acc.\ @ other & difference \\
\midrule
GSM8K      & 0.349 & 87.4 & 85.4 & $+2.0\%$ \\
MultiArith & 0.323 & 99.0 & 98.8 & $+0.2\%$ \\
SVAMP      & 0.371 & 91.6 & 92.8 & $-1.2\%$ \\
AQuA       & 0.358 & 60.4 & 63.8 & $\mathbf{-3.4\%}$ \\
\bottomrule
\end{tabular}
\end{center}
\vspace{-6mm}
\end{table}

We write \emph{crew} for the multiset of roles in a topology, ignoring wiring. We measure concentration as the top-1 topology share, the fraction of questions on which a designer emits its single most frequent shape. Relative to the deployed starting checkpoint, SFT increases the top-1 share from $0.171$ to $0.346$, whereas \method{} reduces it to $0.157$. Resampling each query eight times and comparing each arm against a permutation null built from its own samples, the query explains $1.1\%$ of the cold start's output variety, $2.0\%$ of SFT's and $5.8\%$ of \method{}'s (Appendix~\ref{app:cond}); all three exceed their nulls, so SFT remains query-conditional, and the difference is one of degree. Nor is concentration a loss of vocabulary: the number of distinct shapes is $22$ at the cold start, $40$ for SFT and $34$ for \method{}. We read these diagnostics as ruling out a degenerate explanation of \method{}'s gain, that it wins by emitting a better constant, rather than as evidence of superiority, which \S\ref{sec:control} and \S\ref{sec:transfer} carry.

Table~\ref{tab:fixedcrew} reports the frequency of the modal role assignment and the accuracy associated with its use. The shape is identical on all four benchmarks measured - three agents, two edges, two of the corpus's four roles absent - and is emitted on roughly a third of questions; its value is not, at $+2.0\%$ on GSM8K and $-3.4\%$ on AQuA. Both differences are small beside the $12.60\%$ SFT concedes to the baseline on AQuA, so the modal crew accounts for part of that shortfall rather than all of it.

\subsection{Difficulty-matched control}
\label{sec:control}

\begin{table}[t]
\caption{Cross-arm control. Questions are partitioned by whether SFT emitted its default role assignment, and both arms are scored on exactly those questions, so difficulty is fixed by construction.}
\label{tab:control}
\begin{center}
\small
\begin{tabular}{llccccc}
\toprule
Benchmark & Group & $n$ & SFT & \method{} & $\Delta$ & McNemar $p$ \\
\midrule
\multirow{2}{*}{GSM8K} & SFT emitted default & 460 & 87.4 & 88.5 & $+1.1\%$ & $0.68$ \\
                       & SFT emitted other   & 859 & 85.4 & 87.8 & $+2.4\%$ & $0.15$ \\
\midrule
\multirow{2}{*}{SVAMP} & SFT emitted default & 371 & 91.6 & 94.6 & $+3.0\%$ & $0.16$ \\
                       & SFT emitted other   & 629 & 92.8 & 92.5 & $-0.3\%$ & $0.92$ \\
\midrule
\multirow{2}{*}{AQuA}  & SFT emitted default & 91  & 60.4 & 79.1 & $\mathbf{+18.7\%}$ & $\mathbf{0.0001}$ \\
                       & SFT emitted other   & 163 & 63.8 & 74.8 & $+11.0\%$ & $0.003$ \\
\bottomrule
\end{tabular}
\end{center}
\vspace{-6mm}
\end{table}

On AQuA, SFT has lower accuracy on questions for which it emits its modal role assignment. We therefore partition each benchmark by whether SFT emitted its default crew and score both arms on exactly those questions, so the two face an identical set and difficulty cannot differ between them. Table~\ref{tab:control} shows an advantage in both groups, largest on the questions where SFT emitted its modal role composition. It does not establish that the crew caused the errors, since crew composition, wiring, graph size and the stochastic execution of the agent runtime all vary together across the two designers, so what the partition isolates is the designer and not the crew.

On AQuA, \method{} improves in both groups: $+18.7\%$ where SFT emitted its modal crew ($79.1\%$ against $60.4\%$, exact McNemar $p = 0.0001$) and $+11.0\%$ elsewhere ($p = 0.003$). The margin is largest on the defaulted questions, but it is not confined to them, so the fixed crew accounts for part of the AQuA gap rather than all of it. The other benchmarks separate the groups far less: GSM8K gains about a point in each ($+1.1\%$ against $+2.4\%$, neither significant at the $5\%$ level) and SVAMP gains $3.0\%$ where SFT defaults against $-0.3\%$ elsewhere, neither significant. On those $91$ AQuA questions \method{} emits $18$ distinct topologies and repeats SFT's default crew on only one question in eight, so the crews that replace it are themselves varied.

\section{Discussion}
\label{sec:evidence}

Two objections stand between \S\ref{sec:control} and our account of it: that the channel works through added parameters and training steps rather than through what it carries, and that the advantage is owed to on-policy correction any teacher would supply. We answer the first with measurements independent of downstream accuracy, the second with a matched pair of Stage-2 runs.

\subsection{Channel information content}
\label{sec:shuffle}

Because the FiLM modulation of Eq.~\ref{eq:film} is zero-initialised, whether the teacher uses $a$ at all is directly measurable. On held-out graphs we compute the paired privilege gain
\begin{equation}
\Delta \;=\; \log P_{\text{teacher}}(G \mid Q, a) \;-\; \log P_{\text{teacher}}(G \mid Q),
\label{eq:gain}
\end{equation}
scoring the same teacher on the same graph with the channel on and off, which cancels graph-difficulty variance; a $\Delta$ within $2\sigma$ of zero is consistent with the teacher ignoring the channel. A positive $\Delta$ indicates that the privileged input improves the teacher’s likelihood on held-out verified graphs; this alone does not establish a downstream benefit after distillation, so we pair it with a shuffle control: a teacher identical in FiLM channel, parameter count, optimiser and budget, except that each query is shown another query's graph. If the gain came from capacity or optimisation, the control would show it too.

Figure~\ref{fig:privilege} reports both runs. The shuffled channel's FiLM norm increases from zero during training - the optimiser genuinely trains it - and its gain never leaves zero, so the real channel's $+2.45$ nats is not an artefact of an untrained control. Panel (B) of Figure~\ref{fig:privilege} carries a second, independent signal: the reported FiLM norm reaches 3.10 with sibling inputs and 1.15 with shuffled inputs when it contains something predictive, and does so from the first epochs, before panel (A) has separated.

\subsection{Content versus procedure}
\label{sec:transfer}

The above establishes that the teacher reads the channel, not that what it reads survives distillation into a student that never sees it. We separate content from procedure with a matched Stage~2: two students from the same initialisation, identical in rollouts, $\alpha$, $\lambda$, KL direction, epochs and learning rate, differing only in which teacher supplies the target. The shuffled teacher is fully trained and simply carries no information about this query, so its student still receives on-policy correction on its own visited states - it is \method{} with the privilege removed and nothing else changed.

\begin{figure}[t]
\centering
\begin{minipage}[t]{0.5\textwidth}
\vspace{0pt}
\centering
\IfFileExists{figures/teacher_ablation.pdf}
{\includegraphics[width=\linewidth]{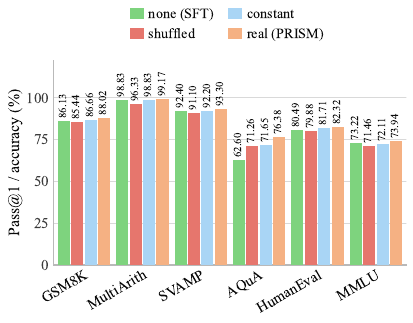}}
{\fbox{\rule[-1.5cm]{0cm}{3.0cm}\rule[-.5cm]{0.9\linewidth}{0cm}}}
\vspace{-6mm}
  \begin{minipage}{0.94\linewidth}
\caption{\method{}'s advantage by Stage-2 teacher: none (SFT), shuffled (another query's graph), constant (privileged input held fixed) or real. All four arms post-train $\theta_0$ under matched budgets, so only the teacher differs.}
\label{fig:transfer}
  \end{minipage}
\end{minipage}\hfill%
\begin{minipage}[t]{0.5\textwidth}
\vspace{0pt}
\centering
\IfFileExists{figures/pi_ablation.pdf}
{\includegraphics[width=\linewidth]{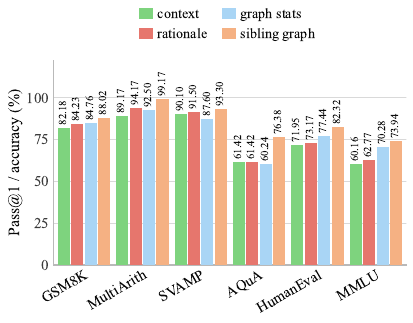}}
{\fbox{\rule[-1.5cm]{0cm}{3.0cm}\rule[-.5cm]{0.9\linewidth}{0cm}}}
\vspace{-6mm}
  \begin{minipage}{0.94\linewidth}
\caption{Privileged channels, with every other Stage-2 setting held fixed. sibling graph, the deployed channel, is best on all six and widest on AQuA; context is weakest on four. The others vary by benchmark.}
\label{fig:pisources}
  \end{minipage}
\end{minipage}
\end{figure}

The constant teacher separates the two effects that the \method{}--SFT gap confounds, and shows that the shuffled teacher is not a clean null: the constant arm beats it on all six benchmarks ($83.86$ against $82.58$ on average), so conditioning on another query's graph misleads the student rather than merely withholding information. \method{} still exceeds the constant teacher on all six ($+1.66\%$ on average, sign test $p = 0.031$) at matched teacher epochs, sampling, supervision, loss mixture and computation, which leaves privileged content as the remaining difference. AQuA carries both contributions, from SFT at $62.60$ to the constant teacher at $71.65$ ($+9.05\%$ from on-policy distillation) and \method{} at $76.38$ ($+4.73\%$ from content), and distillation alone is not sufficient: the constant teacher improves on SFT on three benchmarks only, losing $1.11\%$ on MMLU.

\section{Ablations}
\label{sec:ablation}

\subsection{Privileged channel}

Figure~\ref{fig:pisources} compares the channels with Stage~2 otherwise unchanged. Beyond the query, the teacher sees the question and solution as one string (context), the worked solution, test suite or correct option (rationale), summary statistics of $Q$'s solved set (graph stats), or another graph that also solved $Q$ (sibling graph, our default training-time channel). sibling graph is best on all six, but its margin over the best competitor varies: $3.26\%$ on GSM8K against $14.96\%$ on AQuA.

\textbf{A measured null.} context yields ${\approx}0$ nats, failing the gate of Eq.~\ref{eq:gain}: a correct solution is on-topic by construction, so appending it barely moves the mean-pooled embedding, $5.8\times$ less than shuffled context does (node-level KL $0.0101$ vs.\ $0.0584$). Distilling from it costs $5.84\%$ on GSM8K and $14.96\%$ on AQuA.

\textbf{Node and edge distillation diagnostics.} Every text-based channel drives the edge divergence below $10^{-4}$ with edge-argmax agreement $1.000$, so the $(1-\alpha)$ share of Eq.~\ref{eq:kl} carries almost no signal. The channel enters at the conditioning vector read by the node GRU (Eq.~\ref{eq:film}) and reaches the edge head only through the node state. graph stats alone drives edge KL from $0.52$ to $0.067$, telling the teacher about wiring rather than roles.

\begin{figure}[t]
\centering
\begin{minipage}[t]{0.5\textwidth}
\vspace{0pt}
\centering
\IfFileExists{figures/privilege_gain.pdf}
{\includegraphics[width=\linewidth]{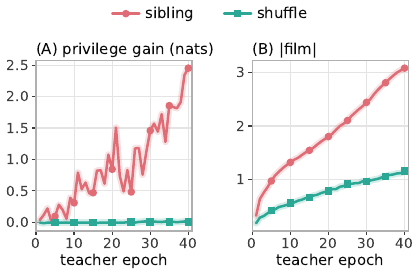}}
{\fbox{\rule[-1.5cm]{0cm}{3.0cm}\rule[-.5cm]{0.9\linewidth}{0cm}}}
\par\vspace{-3pt}
  \begin{minipage}{0.94\linewidth}
\caption{The shuffle control: same channel and budget, different content, the shuffled teacher reading another query's sibling graph. \textbf{(A)} Privilege gain (Eq.~\ref{eq:gain}), on held-out graphs, climbs to $+2.45$ nats for the real channel and stays at zero when shuffled. \textbf{(B)} $\lVert\text{film}\rVert$, zero at initialisation, rises for both, to $3.10$ and $1.15$, so (A) is a null, not a failure to train.}
\label{fig:privilege}
  \end{minipage}
\end{minipage}\hfill%
\begin{minipage}[t]{0.5\textwidth}
\vspace{0pt}
\centering
\IfFileExists{figures/kl_direction.pdf}
{\includegraphics[width=\linewidth]{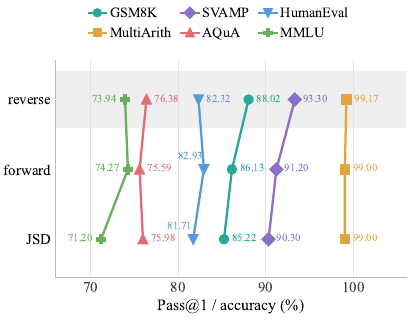}}
{\fbox{\rule[-1.5cm]{0cm}{3.0cm}\rule[-.5cm]{0.9\linewidth}{0cm}}}
\par\vspace{-6pt}
  \begin{minipage}{0.94\linewidth}
\caption{Divergence objectives, with the privileged channel fixed to sibling graph. Reverse KL (our default, shaded) is best on average and on every mathematical benchmark; forward KL and JSD ($\beta = 0.5$) trail it, edging ahead only on HumanEval and MMLU.}
\label{fig:kldir}
  \end{minipage}
\end{minipage}
\end{figure}

\subsection{KL direction}

Figure~\ref{fig:kldir} varies the divergence with the channel fixed to sibling graph: reverse is $\mathrm{KL}(P_\theta \Vert P_{\text{teacher}})$, forward KL reverses the arguments, and JSD their mixture at $\beta = 0.5$. Reverse KL achieves the highest average score among the evaluated divergence objectives ($85.52$ against $84.85$ and $83.90$) and on every mathematical benchmark; forward KL achieves slightly higher accuracy on HumanEval and MMLU by $0.61\%$ and $0.33\%$, gaps small enough that we do not infer a mechanistic explanation from these small differences. The margin is widest on GSM8K and SVAMP, where reverse KL leads forward KL by $1.89\%$ and $2.10\%$ and JSD by $2.80\%$ and $3.00\%$. This is the pattern the mode-seeking argument predicts: with several verified graphs per query, a mass-covering objective spreads the student across the teacher's modes, reproducing the averaging that limits SFT, whereas reverse KL commits it to one.

\section{Conclusion}

We introduced \method{}, a method for post-training autoregressive multi-agent topology generators using successful sibling graphs as privileged inputs. The teacher provides decision-level targets on student-generated trajectories, and the deployed student uses only the query. Under the evaluated Qwen2.5-7B-Instruct configuration, \method{} improves mean accuracy over ARG-Designer across six benchmarks with a similar average number of agent calls. A constant-input teacher control indicates that sibling information contributes beyond the matched distillation procedure. The results do not establish that \method{} recovers a ranking among successful graphs or that query-specific crew assignment explains the gains. Future evaluation should test that assignment effect on representative held-out questions, compare post-training methods at matched compute, and examine transfer across different backbones.

\bibliography{iclr2027_conference}
\bibliographystyle{iclr2027_conference}

\clearpage
\appendix
\section{Hyperparameters}
\label{app:hparams}

Table~\ref{tab:hparams} lists every hyperparameter used for every arm reported in the main text. Values are shared across benchmarks; no per-benchmark tuning was performed.

\begin{table}[h]
\caption{Full hyperparameter listing.}
\label{tab:hparams}
\begin{center}
\small
\begin{tabular}{lll}
\toprule
Stage & Parameter & Value \\
\midrule
\multirow{3}{*}{Shared}
 & sentence encoder & \texttt{all-MiniLM-L6-v2}, $d=384$ \\
 & node/edge loss weight $\alpha$ & $0.2$ \\
 & gradient clipping & $1.0$ \\
\midrule
\multirow{4}{*}{Stage 1 (teacher)}
 & epochs & $40$ \\
 & learning rate & $1\times10^{-4}$ \\
 & privileged rank $r$ & $16$ \\
 & privileged dropout $p$ & $0.1$ \\
\midrule
\multirow{6}{*}{Stage 2 (\method{})}
 & epochs & $30$ \\
 & learning rate & $1\times10^{-4}$ \\
 & queries per step & $16$ \\
 & rollouts per query $M$ & $4$ \\
 & rollout temperature $\tau$ & $1.0$ \\
 & anchor weight $\lambda$ & $0.3$ \\
 & KL direction & reverse \\
 & teacher temperature & $1.0$ \\
\midrule
\multirow{3}{*}{Evaluation}
 & backbone & Qwen2.5-7B-Instruct \\
 & communication rounds $K$ & $1$ \\
 & validation fraction & $0.2$ \\
\bottomrule
\end{tabular}
\end{center}
\end{table}

\section{\method{} training loop}
\label{app:algo}

\begin{algorithm}[h]
\caption{\method{} Stage 2}
\label{alg:opsd}
\begin{algorithmic}[1]
\REQUIRE student $\theta$, frozen privileged teacher $\phi$, query pool $\mathcal{Q}$, anchor loader $\mathcal{A}$
\FORALL{$Q \in \mathcal{Q}$}
  \STATE $a_Q \sim \mathrm{Uniform}(\text{distinct verified topologies of } Q)$ \hfill\COMMENT{kept fixed}
\ENDFOR
\FOR{each optimisation step, with minibatch $B \subset \mathcal{Q}$}
  \FORALL{$Q \in B$}
    \STATE $G^{(1)}, \dots, G^{(M)} \sim P_\theta(\cdot \mid Q)$ \hfill\COMMENT{on-policy, no repairs}
    \FOR{$m = 1, \dots, M$}
      \STATE score $G^{(m)}$ with teacher $\phi(\cdot \mid Q, a_Q)$ \hfill\COMMENT{same $a_Q$ for all $M$ rollouts}
      \STATE score $G^{(m)}$ with student $\theta(\cdot \mid Q)$
    \ENDFOR
  \ENDFOR
  \STATE $G_{\text{real}} \leftarrow$ next batch from $\mathcal{A}$
  \STATE $\mathcal{L} \leftarrow (1-\lambda)\bigl[\alpha\,\mathrm{KL}_{\text{node}} + (1-\alpha)\,\mathrm{KL}_{\text{edge}}\bigr] + \lambda\bigl(-\log P_\theta(G_{\text{real}} \mid Q)\bigr)$
  \STATE update $\theta$ by Adam on $\nabla_\theta \mathcal{L}$ (gradient clipping at $1.0$)
\ENDFOR
\end{algorithmic}
\end{algorithm}

Trajectories are re-sampled at every optimisation step, so the state distribution the teacher corrects tracks the student as it changes. Repairs are applied at evaluation only, never to a sampled graph before scoring. Four edits make a sampled graph executable. First, during evaluation-time generation, any node after the first whose edge head emitted no incoming edge receives one at a uniformly random admissible position, which guarantees $|\mathcal{E}| \geq |\mathcal{V}| - 1$. The generated graph is then made acyclic by deleting one edge from each directed cycle, any disconnected component is joined to the largest by a single edge oriented from the lower to the higher node index, and any node still isolated is linked to a uniformly random other node.

\section{Corpus statistics}
\label{app:corpus}

\textsc{MixChain} contains $2{,}947$ execution-verified graphs over $1{,}014$ distinct queries and $25$ distinct topologies, drawn from five blueprint families (chain, star, layered, fully-connected, mesh) with random role assignments. Mean pairwise Jaccard overlap of the families' solved-query sets is $0.300$; sparse families overlap one another at $0.61$--$0.66$, dense families form a second cluster, and the two clusters are largely disjoint (\S\ref{sec:premise}).

Table~\ref{tab:families} breaks the corpus down by blueprint family, and two features of it carry claims made in \S\ref{sec:collapse}. Across the four families whose attempts were recorded, the solve rates span $58.5$--$59.5\%$, a spread of $1\%$ across families that differ in edge density by a factor of two, so no family is preferred in the aggregate. The corpus shares are not flat, however: they track how often each family was sampled rather than how often it succeeded. That is the combination which leaves the marginal uninformative while the conditional is not.

\begin{table}[h]
\caption{\textsc{MixChain} by blueprint family. cfgs is the number of distinct agent-count configurations instantiated for the family; attempts and solved count executions and successes, and rate is their ratio. corpus is the number of verified graphs the family contributes and share its fraction of the corpus; $|\mathcal{V}|$, $|\mathcal{E}|$ and $|\mathcal{E}|/|\mathcal{V}|$ are means over those graphs. Cold start retains one graph per (query, topology) pair, so the corpus is smaller than the number of successful executions: the $5{,}836$ successes reduce to $2{,}947$ distinct pairs. $^\dagger$Not available for the layered family, so the Total row's attempts, solved and rate cover the other four families.}
\label{tab:families}
\begin{center}
\small
\begin{tabular}{lrrrrrrrrr}
\toprule
Family & cfgs & attempts & solved & rate & corpus & share & $|\mathcal{V}|$ & $|\mathcal{E}|$ & $|\mathcal{E}|/|\mathcal{V}|$ \\
\midrule
Chain           & 2 & $3{,}465$ & $2{,}045$ & $59.0\%$ & $1{,}034$ & $35.1\%$ & $3.03$ & $2.03$ & $0.67$ \\
Star            & 2 & $2{,}872$ & $1{,}681$ & $58.5\%$ & $867$     & $29.4\%$ & $3.63$ & $2.63$ & $0.72$ \\
Layered         & 1 & ---$^\dagger$ & ---$^\dagger$ & ---$^\dagger$ & $541$ & $18.4\%$ & $4.00$ & $4.00$ & $1.00$ \\
Fully-connected & 2 & $2{,}071$ & $1{,}233$ & $59.5\%$ & $286$     & $9.7\%$  & $3.00$ & $3.00$ & $1.00$ \\
Mesh            & 1 & $1{,}494$ & $877$     & $58.7\%$ & $219$     & $7.4\%$  & $4.00$ & $6.00$ & $1.50$ \\
\midrule
Total           & 8 & $9{,}902$ & $5{,}836$ & $58.9\%$ & $2{,}947$ & $100.0\%$ & --- & --- & --- \\
\bottomrule
\end{tabular}
\end{center}
\end{table}

Table~\ref{tab:families_mbpp} gives the same breakdown for the code-generation corpus. The MBPP cold start used the default configurations of \citet{li2026argdesigner} for code generation, which are dense-only, so there is no chain, star or layered family to report: the corpus is built from fully-connected and mesh blueprints alone. The pattern of Table~\ref{tab:families} survives the change of domain. The two families solve at $63.6\%$ and $62.4\%$, a spread of $1.2\%$ despite differing in edge density, while their corpus shares, $65.5\%$ against $34.5\%$, again follow how often each was instantiated. Cold start's one-graph-per-(query, topology) rule reduces $1{,}259$ successes to $981$ graphs.

\begin{table}[h]
\caption{MBPP by blueprint family, with columns as in Table~\ref{tab:families}. Only dense families appear because the code-generation cold start follows the default configuration of \citet{li2026argdesigner}. The $1{,}259$ successful executions reduce to $981$ distinct (query, topology) pairs.}
\label{tab:families_mbpp}
\begin{center}
\small
\begin{tabular}{lrrrrrrrrr}
\toprule
Family & cfgs & attempts & solved & rate & corpus & share & $|\mathcal{V}|$ & $|\mathcal{E}|$ & $|\mathcal{E}|/|\mathcal{V}|$ \\
\midrule
Fully-connected & 3 & $1{,}312$ & $835$     & $63.6\%$ & $643$ & $65.5\%$  & $4.04$ & $6.46$ & $1.60$ \\
Mesh            & 2 & $679$     & $424$     & $62.4\%$ & $338$ & $34.5\%$  & $4.37$ & $7.49$ & $1.71$ \\
\midrule
Total           & 5 & $1{,}991$ & $1{,}259$ & $63.2\%$ & $981$ & $100.0\%$ & ---    & ---    & --- \\
\bottomrule
\end{tabular}
\end{center}
\end{table}

\section{Seed variability}
\label{app:seeds}

Every arm is post-trained three times from the same deployed checkpoint $\theta_0$ under seeds $42$, $43$ and $44$, the runs differing in nothing else, and each run is evaluated on the full benchmark suite. Table~\ref{tab:seeds} reports the mean and sample standard deviation over the three runs; seed $42$ is the run reported in Table~\ref{tab:main} and everywhere else in the paper.

\begin{table}[h]
\caption{Accuracy over three post-training runs per arm, under seeds $42$, $43$ and $44$. Each cell is the mean, with the sample standard deviation over the three runs as a subscript.}
\label{tab:seeds}
\begin{center}
\small
\setlength{\tabcolsep}{4pt}
\begin{tabular}{lcccccc}
\toprule
Arm & GSM8K & MultiArith & SVAMP & AQuA & HumanEval & MMLU \\
\midrule
\argd{} & $80.21_{\pm 0.75}$ & $98.33_{\pm 0.50}$ & $92.93_{\pm 0.51}$ & $75.33_{\pm 1.38}$ & $79.07_{\pm 0.35}$ & $72.28_{\pm 0.64}$ \\
SFT & $86.18_{\pm 0.09}$ & $98.83_{\pm 0.00}$ & $92.73_{\pm 0.42}$ & $62.73_{\pm 0.61}$ & $80.29_{\pm 0.35}$ & $73.09_{\pm 0.54}$ \\
constant teacher & $86.71_{\pm 0.04}$ & $98.66_{\pm 0.29}$ & $92.37_{\pm 0.29}$ & $69.42_{\pm 1.94}$ & $81.51_{\pm 0.35}$ & $73.29_{\pm 1.02}$ \\

\method{} & $\mathbf{87.77}_{\pm 0.57}$ & $\mathbf{98.89}_{\pm 0.26}$ & $\mathbf{93.47}_{\pm 0.47}$ & $\mathbf{76.51}_{\pm 1.38}$ & $\mathbf{82.52}_{\pm 0.35}$ & $\mathbf{73.76}_{\pm 0.55}$  \\
\bottomrule
\end{tabular}
\end{center}
\end{table}

Run-to-run spread is small beside the differences the paper reports. The largest standard deviation on any benchmark is $1.94$, on AQuA, where the three arms conditioned on the query alone are least stable, and most lie below $0.6$, while the benchmark-averaged accuracy varies by at most $0.20$ within an arm. Against that, \method{} leads SFT by $3.18$ on the benchmark average, an order of magnitude larger than either arm's spread.

The per-benchmark picture is the one the main text describes, with one qualification. Treating the seeds as paired runs, \method{} exceeds SFT in all three on GSM8K, SVAMP, AQuA, HumanEval and MMLU; on MultiArith the two are indistinguishable ($98.89$ against $98.83$, paired difference $+0.06 \pm 0.26$, with \method{} ahead in one seed, level in one and behind in one), so the $0.34\%$ margin visible in Table~\ref{tab:main} is seed noise rather than an effect, and MultiArith is better read as saturated. \method{} exceeds \argd{} on the mean of every benchmark, including MultiArith ($+0.56$) and AQuA ($+1.18$), and does so in all three seeds.

Two properties of SFT also reproduce. Its AQuA accuracy remains near 63\% across the evaluated seeds while \argd{} stays near $75$, so the uneven transfer documented in \S\ref{sec:collapse} is a property of the training configuration and not an unlucky run. And SFT's benchmark average, $82.31_{\pm 0.10}$, sits below the checkpoint it starts from, $83.03_{\pm 0.17}$: continued supervised fine-tuning on the verified corpus buys large gains on GSM8K and HumanEval and pays for them elsewhere, which is the pattern \method{} is designed to avoid.

\section{FiLM rank}

\noindent
\begin{minipage}[t]{0.50\textwidth}
\vspace{0pt}
Figure~\ref{fig:rank} varies the rank from the default $r = 16$ to a full-width channel, a ${\sim}30\times$ range in channel parameters. Every setting stays within $2.7\%$ of the default, less than \method{}'s own $3.2\%$ average margin over SFT; the largest deviation is MMLU ($-2.7\%$) at $r = 64$. The ordering inside that range is not systematic: $r = 64$ is weakest on three of the six benchmarks, and one setting edges past the default on two benchmarks while another matches it. The four rank averages lie within $1.49\%$, all above SFT ($82.28$), and accuracy does not track capacity; rank is not a sensitive hyperparameter here.
\end{minipage}\hfill%
\begin{minipage}[t]{0.47\textwidth}
\vspace{-3pt}
\centering
\IfFileExists{figures/rank_ablation.pdf}
  {\includegraphics[width=\linewidth]{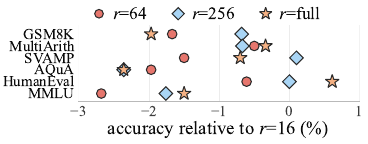}}
  {\fbox{\rule[-1.5cm]{0cm}{3.0cm}\rule[-.5cm]{0.9\linewidth}{0cm}}}
\vspace{-11pt}
\captionof{figure}{Accuracy relative to the default rank $r = 16$ (Eq.~\ref{eq:film}), all else fixed.}
\label{fig:rank}
\end{minipage}

\section{Corpus splits and sibling availability}
\label{app:splits}

Table~\ref{tab:splits} reports the three verified corpora of \S\ref{sec:experiments}, their query-level splits, and the share of queries for which a privileged sibling exists.

\begin{table}[h]
\caption{Verified corpora, splits, and sibling availability. Queries with a sibling is the share having at least two distinct verified topologies; the remainder receive a zero privileged vector, so the teacher is conditioned on the query alone. Validation is a query-level $20\%$ of each corpus, used only to select the teacher checkpoint; it is not drawn from any evaluation set.}
\label{tab:splits}
\begin{center}
\small
\setlength{\tabcolsep}{4pt}
\begin{tabular}{lrrrrrcc}
\toprule
Corpus & graphs & queries & \shortstack{graphs\\per query} & \shortstack{topologies\\per query} & \shortstack{with\\sibling} & \shortstack{train / val\\queries} & \shortstack{train / val\\graphs} \\
\midrule
\textsc{MixChain} & $2{,}947$ & $1{,}014$ & $2.91$ & $2.91$ & $76.9\%$ & $811$ / $203$ & $2{,}359$ / $588$ \\
MMLU              & $1{,}890$ & $249$     & $7.59$ & $7.56$ & $87.1\%$ & $199$ / $50$  & $1{,}524$ / $366$ \\
MBPP              & $981$     & $269$     & $3.65$ & $3.65$ & $90.3\%$ & $215$ / $54$  & $777$ / $204$ \\
\bottomrule
\end{tabular}
\end{center}
\end{table}

Graphs per query and distinct topologies per query are all but identical in every corpus ($2.91$ against $2.91$, $7.59$ against $7.56$, $3.65$ against $3.65$), which fixes how the cold start treats repeated successes: it retains at most one graph per (query, topology) pair. That accounts for the gap between execution and corpus counts in Table~\ref{tab:families}, where $5{,}836$ successes reduce to $2{,}947$ graphs, each distinct pair having been solved roughly twice on average.

Sibling availability is high but not universal: $23.1\%$ of \textsc{MixChain} queries, $12.9\%$ of MMLU queries, and $9.7\%$ of MBPP queries have a single verified topology and therefore fall back to query-only conditioning (\S\ref{sec:method}). This bounds the fraction of each training set the privileged channel can reach, and it is a corpus property the ablations can be read against: coverage is widest on MBPP and MMLU, and \textsc{MixChain}, the largest corpus, is also the one whose queries most often admit a single verified topology.

\section{Output conditionality}
\label{app:cond}

Each designer is resampled eight times per query on a $200$-question GSM8K subset at $\tau = 1.0$, a topology being (role multiset, $|\mathcal{V}|$, $|\mathcal{E}|$). $H(G)$ is the entropy of the pooled shape distribution, $H(G \mid Q)$ the mean entropy within a query, and $I(Q;G) = H(G) - H(G\mid Q)$.

\begin{table}[h]
\caption{Output conditionality. \emph{null} is the mean $I(Q;G)$ the same arm produces when its own samples are reassigned to random queries, i.e. its estimator bias at this support and sample count; that bias grows with the number of distinct outcomes and so is not equal across arms, which is why each arm is judged against its own null rather than a shared correction. \emph{excess} is $I$ minus that null. $p$ is a one-sided permutation test over $200$ draws; the interval is a percentile bootstrap over queries ($500$ resamples). \textbf{Bold} marks the larger excess of the two arms post-trained from $\theta_0$, SFT and \method{}.}
\label{tab:cond}
\begin{center}\small\setlength{\tabcolsep}{3.4pt}
\begin{tabular}{lcccccccc}
\toprule
Arm & $H(G)$ & $H(G\mid Q)$ & $I$ & null & excess & $p$ & 95\% CI & top-1 \\
\midrule
cold start & $1.569$ & $1.041$ & $0.528$ & $0.511$ & $0.017$ & $0.045$ & $[0.485,\, 0.558]$ & $0.557$ \\
SFT                       & $2.584$ & $1.497$ & $1.087$ & $1.036$ & $0.051$ & $0.005$ & $[1.031,\, 1.116]$ & $0.346$ \\
\method{}                 & $2.772$ & $1.584$ & $1.188$ & $1.028$ & $\mathbf{0.160}$ & $0.005$ & $[1.124,\, 1.221]$ & $0.157$ \\
\argd{}            & $2.336$ & $1.227$ & $1.109$ & $0.732$ & $0.377$ & $0.005$ & $[1.047,\, 1.152]$ & $0.171$ \\
\bottomrule
\end{tabular}
\end{center}
\end{table}

All three arms that share the cold-start initialisation show query dependence above their own null (permutation $p \leq 0.045$), so we do not claim that supervised fine-tuning ignores the query. The sizes differ substantially: the excess of $I(Q;G)$ over each arm's null is $0.017$ nats for the cold start, $0.051$ for SFT and $0.160$ for \method{}, which is $1.1\%$, $2.0\%$ and $5.8\%$ of each arm's output entropy, and the bootstrap intervals for SFT and \method{} do not overlap. Most of the raw plug-in statistic is estimator bias at eight samples per query, which is why the excess rather than the raw value is the comparable quantity. The deployed \argd{} checkpoint is the most query-conditional arm measured (excess $0.377$ nats over its own null) while being the least accurate of the three deployed designers in Table~\ref{tab:main}. Greater measured query dependence therefore does not necessarily correspond to higher downstream accuracy. These diagnostics show that the generated graph distributions depend on the query; they do not establish that this dependence accounts for the accuracy improvement.

\section{Topology intervention}
\label{app:intervention}

The controls of \S\ref{sec:control} compare designers, so topology and the designer that produced it vary together. We therefore execute chosen crews directly. The question set is the $91$ AQuA questions on which SFT emitted its modal crew -- the ``SFT emitted default'' group of Table~\ref{tab:control}. Every condition supplies only a role multiset; wiring is a chain in all conditions, with one round, the same decision method and the same backbone, and each condition is executed three times.

\begin{table}[h]
\caption{Executing chosen crews on the same $91$ AQuA questions. \emph{mean} is the average accuracy over three executions and \emph{vote} the accuracy of the per-question majority verdict across them; the McNemar counts in Table~\ref{tab:intervention-pairs} are computed from that verdict and pair with \emph{vote}, not with \emph{mean}. SFT's own crew and its modal crew in canonical order share a role multiset by construction and differ only in the order of roles along the chain.}
\label{tab:intervention}
\begin{center}\small
\begin{tabular}{lccccc}
\toprule
Condition & mean & sd & min--max & vote & correct \\
\midrule
SFT's own crew                              & $61.54$ & $1.79$ & $59.34$--$63.74$ & $60.44$ & $55/91$ \\
SFT's modal crew, canonical order           & $53.85$ & $2.37$ & $51.65$--$57.14$ & $53.85$ & $49/91$ \\
size-matched random crew                    & $75.46$ & $0.52$ & $74.73$--$75.82$ & $82.42$ & $75/91$ \\
\method{}'s crews, permuted across questions& $78.39$ & $2.07$ & $76.92$--$81.32$ & $79.12$ & $72/91$ \\
\method{}'s own crew                        & $77.66$ & $4.14$ & $74.73$--$83.52$ & $79.12$ & $72/91$ \\
\bottomrule
\end{tabular}
\end{center}
\end{table}

\begin{table}[h]
\caption{Paired comparisons on the same $91$ questions. $b$ and $c$ are questions solved only by the first and only by the second condition; exact McNemar uses those two counts alone. $p$(vote) is computed on the majority verdict across three executions; $p$ per run tests each execution separately.}
\label{tab:intervention-pairs}
\begin{center}\small
\begin{tabular}{lcccl}
\toprule
Comparison & $b$ & $c$ & $p$(vote) & $p$ per run \\
\midrule
modal crew vs \method{}        & $7$ & $30$ & $\mathbf{0.0002}$ & $0.002,\ 0.001,\ 0.000$ \\
SFT's crew vs \method{}        & $4$ & $21$ & $\mathbf{0.0009}$ & $0.087,\ 0.036,\ 0.000$ \\
modal crew vs random           & $2$ & $28$ & $\mathbf{<0.0001}$ & $0.003,\ 0.001,\ 0.000$ \\
\method{} vs random            & $6$ & $9$  & $0.6072$ & $1.000,\ 1.000,\ 0.118$ \\
\method{} vs permuted          & $9$ & $9$  & $1.0000$ & $0.832,\ 0.832,\ 0.804$ \\
permuted vs random             & $4$ & $7$  & $0.5488$ & $1.000,\ 0.839,\ 0.302$ \\
\bottomrule
\end{tabular}
\end{center}
\end{table}

Executing SFT's own crew reaches $61.54\%$ against \method{}'s $77.66\%$ on the same questions, a difference of $4$ against $21$ paired disagreements (exact McNemar $p = 0.0009$). With the designers removed from the comparison and wiring, rounds, decision method and backbone held fixed, the crew is what differs, so this is causal evidence at the level of topology rather than the association of \S\ref{sec:control}.

Two further conditions bound how much this supports. A size-matched random crew reaches $75.46\%$ and is not distinguishable from \method{}'s crews ($6$ against $9$, $p = 0.61$), and we detect no statistically significant difference between the original and permuted crew assignments on this subset ($9$ against $9$, $p = 1.00$). The supported conclusion is therefore narrow: SFT's modal crew is worse than the alternatives on these questions, and both the random-crew and permuted-crew controls recover much of the observed performance gap. This intervention does not establish that \method{}'s particular crews, or their assignment to particular queries, carry benefit beyond that. We note that the question set is selected by SFT's behaviour, which works against detecting a matching effect, so the permutation result is a failure to detect rather than evidence of absence.

Composition is also not the whole story. Forcing the modal crew in canonical order scores $53.85\%$ against $61.54\%$ for the same multiset in the order SFT emitted, a difference larger than any per-condition spread in Table~\ref{tab:intervention}, so role order along the chain matters at fixed composition and the crew abstraction used throughout \S\ref{sec:collapse} is not order-invariant.

\section{Evaluation benchmarks}
\label{app:benchmarks}

Table~\ref{tab:benchmarks} lists the six benchmarks on which every arm is evaluated.

\begin{table}[h]
\caption{Benchmarks used for evaluation.}
\label{tab:benchmarks}
\begin{center}
\small
\begin{tabular}{llcr}
\toprule
Category & Dataset & Answer Type & \#Test \\
\midrule
General reasoning & MMLU & Multi-choice & $1{,}531$ \\
\midrule
\multirow{4}{*}{Math reasoning}
 & GSM8K & Number & $1{,}319$ \\
 & MultiArith & Number & $600$ \\
 & SVAMP & Number & $1{,}000$ \\
 & AQuA & Multi-choice & $254$ \\
\midrule
Code generation & HumanEval & Code & $164$ \\
\bottomrule
\end{tabular}
\end{center}
\end{table}

\FloatBarrier

\section{Qualitative Examples}
\label{app:examples}

To make the difference between designers concrete, we show the crew each one builds, and the answer the resulting system returns, on six GSM8K problems (Qwen2.5-7B-Instruct). In each, the answer shown is the one the system states explicitly, so no answer parsing is involved.

\needspace{19\baselineskip}
\subsection{Example 1: Problem 652}

An arithmetic word problem. The baseline routes it through two Programming Experts and an Inspector and returns $105$; \method{} assigns three Math Solvers and reaches the gold answer of $75$.

\begin{solbox}[Problem \hfill Gold answer: 75]{cneu}
James hires a horse-drawn carriage from 5 PM to 9 PM. He gets 1 hour free. The first paid hour is \$15 and each hour after that is twice the cost. How much did he pay?
\end{solbox}

\begin{solbox}[\argd{} (baseline) \hfill Answer: 105 ~$\times$]{cbad}
\textbf{Crew:} Programming Expert $\rightarrow$ Inspector $\rightarrow$ Programming Expert
\end{solbox}

\begin{solbox}[\method{} \hfill Answer: 75 ~$\checkmark$]{cok}
\textbf{Crew:} Math Solver $\rightarrow$ Math Solver $\rightarrow$ Math Solver
\end{solbox}

\subsection{Example 2: Problem 847}

A simple-interest problem. The baseline assigns three Inspectors and returns $10{,}298.44$, an order of magnitude above the gold answer; \method{} replaces the middle Inspector with a Mathematical Analyst and returns $975$.

\begin{solbox}[Problem \hfill Gold answer: 975]{cneu}
Brenda invested \$300 into a scheme that accumulates simple interest at the rate of three-quarters of the original amount per year. How much in total will she have after 3 years?
\end{solbox}

\begin{solbox}[\argd{} (baseline) \hfill Answer: 10298.44 ~$\times$]{cbad}
\textbf{Crew:} Inspector $\rightarrow$ Inspector $\rightarrow$ Inspector
\end{solbox}

\begin{solbox}[\method{} \hfill Answer: 975 ~$\checkmark$]{cok}
\textbf{Crew:} Inspector $\rightarrow$ Mathematical Analyst $\rightarrow$ Inspector
\end{solbox}

\needspace{19\baselineskip}
\subsection{Example 3: Problem 1312}

A multi-step revenue problem. The baseline assigns three Inspectors, and its logged reasoning drops the six reservations entirely: it prices a single reservation at \$15 a night and returns $30$. \method{} assigns three Math Solvers, which keep every step ($6 \times \$15 = \$90$ a night, then $\$90 \times 2 = \$180$) and reach the gold answer.

\begin{solbox}[Problem \hfill Gold answer: 180]{cneu}
Tom's restaurant gets 6 reservations a night. They normally order 2 meals that cost \$5 each and a \$5 bottle of wine. How much do they make a week if they are open 2 days a week?
\end{solbox}

\begin{solbox}[\argd{} (baseline) \hfill Answer: 30 ~$\times$]{cbad}
\textbf{Crew:} Inspector $\rightarrow$ Inspector $\rightarrow$ Inspector
\end{solbox}

\begin{solbox}[\method{} \hfill Answer: 180 ~$\checkmark$]{cok}
\textbf{Crew:} Math Solver $\rightarrow$ Math Solver $\rightarrow$ Math Solver
\end{solbox}

\needspace{19\baselineskip}
\subsection{Example 4: Problem 763}

The baseline, with two Programming Experts and an Inspector, sets up the wrong equation ($x + x/2 + 2x = 250$), obtains $142.86$ and rounds it to $143$, although a count of cases must come out whole. \method{} assigns two Math Solvers and recovers $50$, $100$ and $100$ cases for the three houses.

\begin{solbox}[Problem \hfill Gold answer: 100]{cneu}
Rob, Royce, and Pedro are contractors getting ready to put a new roof on three homes. If the three homes will need 250 cases of shingles, with the first house needing 1/2 of the second, and the third needing double the first. How many cases of shingles will the third house need?
\end{solbox}

\begin{solbox}[\argd{} (baseline) \hfill Answer: 143 ~$\times$]{cbad}
\textbf{Crew:} Programming Expert $\rightarrow$ Inspector $\rightarrow$ Programming Expert
\end{solbox}

\begin{solbox}[\method{} \hfill Answer: 100 ~$\checkmark$]{cok}
\textbf{Crew:} Math Solver $\rightarrow$ Math Solver
\end{solbox}

\subsection{Example 5: Problem 322}

The baseline routes the problem through two Programming Experts and an Inspector, the same role composition it builds in Examples~1 and~4, and returns $8$. \method{} assigns an Inspector, a Mathematical Analyst and a Math Solver and reaches the gold answer of $9$.

\begin{solbox}[Problem \hfill Gold answer: 9]{cneu}
Jam has three boxes full of pencils and 2 loose pencils which give a total of 26 pencils. If her sister, Meg, has 46 pencils, how many boxes do Jam and Meg need to store all their pencils?
\end{solbox}

\begin{solbox}[\argd{} (baseline) \hfill Answer: 8 ~$\times$]{cbad}
\textbf{Crew:} Programming Expert $\rightarrow$ Inspector $\rightarrow$ Programming Expert
\end{solbox}

\begin{solbox}[\method{} \hfill Answer: 9 ~$\checkmark$]{cok}
\textbf{Crew:} Inspector $\rightarrow$ Mathematical Analyst $\rightarrow$ Math Solver
\end{solbox}

\needspace{19\baselineskip}
\subsection{Example 6: Problem 664}

The baseline assigns three Inspectors, the same crew it builds in Examples~2 and~3, and returns $48$, which is the total class time in hours rather than the number of recipes. \method{} assigns an Inspector, a Programming Expert and a Math Solver and reaches the gold answer of $32$.

\begin{solbox}[Problem \hfill Gold answer: 32]{cneu}
John decides to take a cooking class. The class meets 4 times a week for 2 hours each time for 6 weeks. If he learns a new recipe for every 1.5 hours of class time, how many recipes does he learn?
\end{solbox}

\begin{solbox}[\argd{} (baseline) \hfill Answer: 48 ~$\times$]{cbad}
\textbf{Crew:} Inspector $\rightarrow$ Inspector $\rightarrow$ Inspector
\end{solbox}

\begin{solbox}[\method{} \hfill Answer: 32 ~$\checkmark$]{cok}
\textbf{Crew:} Inspector $\rightarrow$ Programming Expert $\rightarrow$ Math Solver
\end{solbox}

\end{document}

%% file: math_commands.tex
\usepackage{amsmath,amsfonts,bm}

\def\eqref#1{equation~\ref{#1}}

\def\1{\bm{1}}

\DeclareMathAlphabet{\mathsfit}{\encodingdefault}{\sfdefault}{m}{sl}
\SetMathAlphabet{\mathsfit}{bold}{\encodingdefault}{\sfdefault}{bx}{n}

